\documentclass{article}
\usepackage{spconf,amsmath,graphicx}
\usepackage{amssymb}
\usepackage{multirow,setspace,verbatim,amsfonts,graphicx,amsmath,amsthm,amsbsy,amssymb,epsfig,url}
\usepackage{amsthm}
\usepackage{amsfonts}
\usepackage{amsmath,bm}
\usepackage{amssymb}
\usepackage{amsthm}
\usepackage{tikz,graphicx}
\usepackage{mathrsfs} 
\usepackage{algpseudocode}

\newcommand{\Rd}{{\mathbb R}}

\newcommand{\hank}{\mathbb{H}}

\title{Deep Learning  for Accelerated Ultrasound Imaging 
}
\name{Yeo Hun Yoon and Jong Chul Ye \sthanks{This work is supported by National Research Foundation of Korea (NRF-2016R1A2B3008104).}}
\address{
Bio Imaging \& Signal Processing Lab., Dept. of Bio and Brain Engineering, KAIST\\
291 Daehak-ro, Yuseong-gu, Daejeon, Republic of Korea}
\begin{document}
%
\maketitle
\begin{abstract}
In  portable,  3-D,  or ultra-fast ultrasound (US) imaging systems,
there is an increasing demand 
to reconstruct high quality images from limited number of data.
%
However, 
the existing solutions
 require either hardware changes or computationally expansive algorithms.
 To overcome these limitations, here we propose a novel deep learning approach
 that interpolates the missing RF data by utilizing  the sparsity of the RF data in the Fourier domain.
Extensive experimental  results from sub-sampled RF data from a real US system
confirmed that the proposed method can effectively reduce the data rate  without sacrificing the image quality.
\end{abstract}
\section{Introduction}
\label{sec:intro}

Due to the the excellent temporal resolution with reasonable image quality and minimal invasiveness, ultrasound imaging  
 has been adopted as a golden-standard for many disease diagnosis in heart, liver, etc.
Accordingly, there have been many research efforts to extend the US imaging to new applications such as portable imaging
in emergency care, 3-D imaging, ultra-fast imaging, etc.
However, the common technical hurdle in these applications is to reconstruct high resolution images
from significantly reduced radio-frequency (RF) data acquired with an US transducer.

While the compressed sensing approach has been investigated to address this issue \cite{liebgott2012pre}, accurate modeling of wave propagation is usually 
required.
Recently,  Wagner et al \cite{wagner2012compressed} models a scan line profile as finite rate of innovation, 
and proposed a specially-designed hardware architecture that enables high resolution scan line reconstruction 
using sparsity driven recovery algorithm \cite{wagner2012compressed}.  
Another recent proposal is to use a  low rank Hankel matrix completion approach \cite{ye2016compressive}. In particular, thanks to the strong correlation between scan lines and adjacent temporal frames, Jin et al \cite{jin2016compressive} showed that
a rank deficient Hankel structured matrix can be obtained from the reordered RF measurement data. Accordingly,
 the missing data can be accurately reconstructed using a low rank Hankel matrix completion algorithm \cite{ye2016compressive}.
However, the algorithm is computationally very expensive, which is not suitable for routine diagnostic applications.

One of the most important contributions of this paper is therefore to show that the low-rank interpolations problem of RF data can be solved using deep convolutional
neural network (CNN), which enables very fast run-time reconstruction. In particular, inspired by
recent finding that a CNN can be interpreted as a {\em deep convolutional framelets} obtained from a decomposition of Hankel matrix  \cite{ye2017deep},
we construct a CNN that performs direct interpolation of RF data. Thus, the final reconstruction can be done using the standard delay-and-sum (DAS) beamformer
without changing any hardware/software structures.
Compared to an image domain CNN that attempts to learn acquisition geometry-dependent artifacts, 
one of the most important advantages of the proposed RF domain CNN is its universality.
%
Specifically, although an image domain deep learning requires  many data set from various
acquisition geometry and body parts \cite{kang2017deep},  
our CNN can be trained using a RF data measured by, for example,  the linear array transducer for a particular organ, which
can be then used for other type of  transducers and/or different body parts. Therefore, the proposed
systems is very practical in real applications.

Extensive experimental results using RF data from a real US system confirmed that the proposed method has significant potential for 
accelerated US systems.

\section{Theory}
\label{sec:thoery}

\begin{figure}[htb]
\begin{minipage}[b]{0.48\linewidth}
  \centering
  \centerline{\includegraphics[width=3.cm]{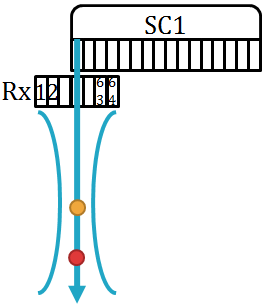}}
  \centerline{(a) Beam emission}\medskip
\end{minipage}
\hfill
\begin{minipage}[b]{0.48\linewidth}
  \centering
  \centerline{\includegraphics[width=3.5cm]{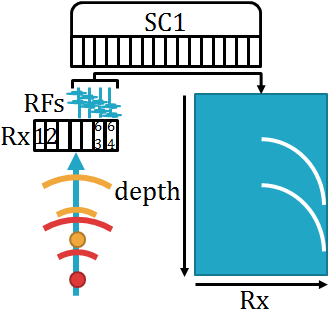}}
  \centerline{(b) RFs record}\medskip
\end{minipage}
\begin{minipage}[b]{1.0\linewidth}
  \centering
  \centerline{\includegraphics[width=7.5cm]{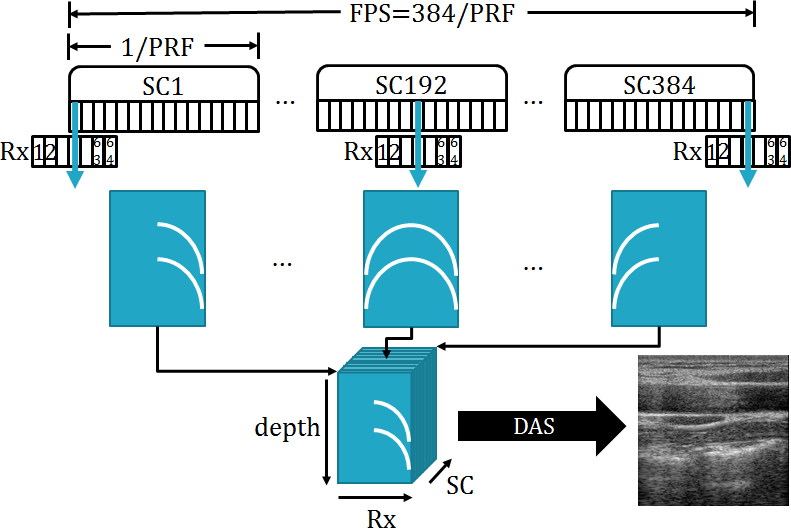}}
  \centerline{(c) conventional data acquisitions}
\end{minipage}
\vspace*{-0.5cm}
\caption{B-mode ultrasound imaging flow.}
\label{fig:theorya}
\end{figure}
%

\subsection{Sub-sampled dynamic aperture US system}

As a proof-of-concept,  we consider a sub-sampled dynamic aperture US system \cite{jin2016compressive}.
Specifically, a 
B-mode ultrasound scans the body and displays the 2-D image as shown in Fig.\ref{fig:theorya}. In Fig.\ref{fig:theorya}, SC, Rx and DAS denote scan line for each ultrasound beam, receivers of the transducer, and delay-and-sum (DAS) beamformer, respectively. After a focused ultrasound beam is illuminated along the scan line as shown in Fig.\ref{fig:theorya}(a), ultrasound beam is reflected at some tissue boundaries and the reflected US data are recorded by receivers as RF data  (see Fig.\ref{fig:theorya}(b)). 
Thus, for each scan line (SC), a  Depth-Rx coordinate RF data  is obtained, and this is repeated for each scan line to obtain
a 3-D cube of RF data in Depth-Rx-SC coordinates.
Then, a DAS beamformer uses the 3-D stacked data to generate one temporal frame as shown in Fig.\ref{fig:theorya}(c). 
Because of huge dat set, a hardware  beamformer is usually incorporated to a transducer and the generated image is  transferred to an image workstation.
On the other hand, recent software-based US systems directly transfer RF data to the image workstation, where  a software-based beamformer reconstructs images.
However, for many portable or high spatio/temporal applications,  online reconstruction or huge data transfer may not be a viable option.

%

\begin{figure*}[!htb]
  \centering
  \centerline{\includegraphics[width=\textwidth]{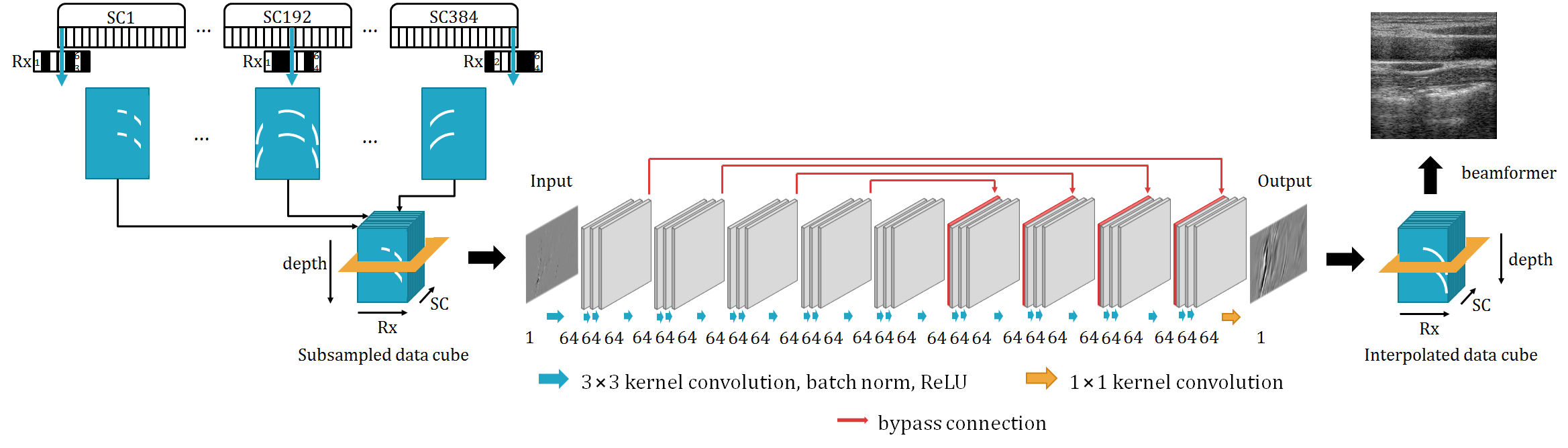}}
\caption{Network architecture for B-mode ultrasound reconstruction}
\label{fig:methodb}
\end{figure*}

In this paper, we thus consider a randomly sub-sampled dynamic aperture US system that  acquires only partial data as shown in Fig.\ref{fig:methodb}.
Because hardware is not changed, this method can be easily used in any conventional ultrasound system. 
Also, it can relieve the frame rate limitation caused by pulse repetition frequency. 

\subsection{Redundancy of RF data and low-rank Hankel matrix}

\begin{figure}[htb]
  \centering
  \centerline{\includegraphics[width=8cm]{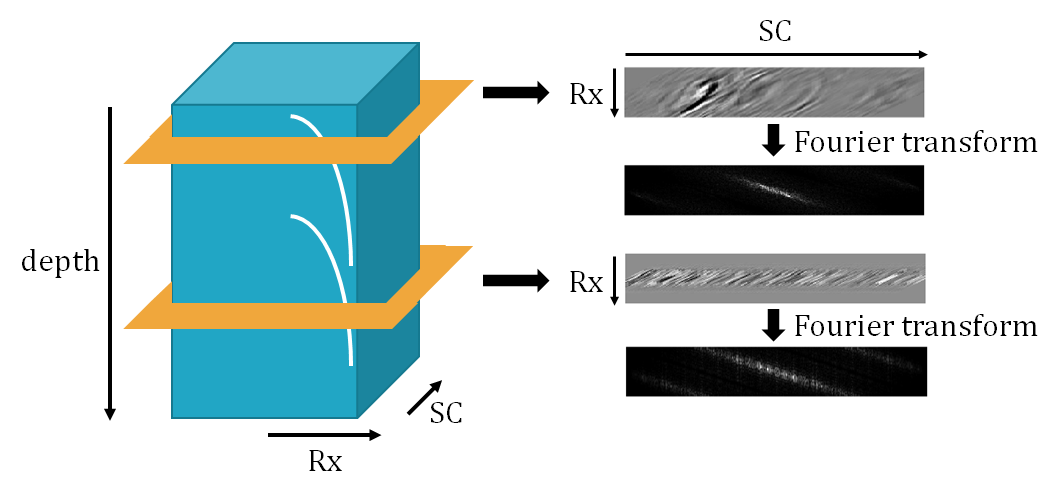}}
\caption{Extraction of Rx-SC data for each depth direction.}
\label{fig:methoda}
\end{figure}

Recall that B-mode ultrasound measurements are acquired by point-by-point scanning for each scan line. 
Since the scan line only changes incrementally, the acquired Rx data along detectors do not change rapidly
for each scan line.
This implies that there exists some level of skewed redundancy in Rx-SC coordinate data, which can be also observed in 
Fig.~\ref{fig:methoda}.
 
This type of redundancy can be easily seen as a sparsity in the Fourier domain as demonstrated in  
Fig.~\ref{fig:methoda}.
Therefore, when the Fourier component of the Rx-SC image is given by $\widehat{f}$, then we can find an annihilating function $\widehat{h}$ in the spectral domain such that their multiplication becomes zero:
\begin{equation}
\widehat{f} ~\widehat{h} =0
\end{equation}
This is equivalent to the convolutional relationship in the Rx-SC  domain:
\begin{eqnarray}
{f} \circledast {h} =0,
\end{eqnarray}
which can be represented in a matrix form :
\begin{eqnarray}
\hank(f) h =0,
\end{eqnarray}
where $\hank(f)$ is a Hankel structured  matrix constructed from $f$. 
This implies that  the Hankel matrix constructed from reordered raw data is rank-deficient.
Furthermore, its  rank is determined by sparsity level as theoretically proven in \cite{ye2016compressive}.
In fact, Jin et al \cite{jin2016compressive} utilized this idea to interpolate missing RF data using low-rank Hankel matrix completion.
However, the main limitation of \cite{jin2016compressive} is its computational complexity, which is prohibitive in real applications.

\subsection{Deep Convolutional Framelets}

Recently, it was shown that CNN is closely related to Hankel matrix decomposition \cite{ye2017deep}.
Specifically, for a given Hankel matrix $\hank(f) \in \Rd^{n\times d}$, let
 $\Phi$ and $\tilde \Phi \in \Rd^{n\times m}$ (resp. $\Psi$ and $\tilde \Psi \in \Rd^{d\times q}$) are frame and its dual frame, respectively, 
satisfying the frame condition:
$ \tilde \Phi \Phi^\top = I,  \Psi \tilde \Psi^{\top} = I$
 such that it satisfies the perfect reconstruction condition:
 \begin{eqnarray}\label{eq:PR}
 \hank(f) = \tilde \Phi \Phi^\top \hank(f) \Psi \tilde\Psi^\top.
 \end{eqnarray}
One of the most important discoveries  in  \cite{ye2017deep} is 
to show that  \eqref{eq:PR} can be equivalently
represented in the signal domain using so-called {\em deep convolutional framelets} expansion, where
 the decoder-part  convolution is given by
\begin{eqnarray}
f  
&=&  \left(\tilde\Phi C\right) \circledast \tau(\tilde \Psi) \label{eq:frameeq}
\end{eqnarray}
where $C$ is  the framelet coefficient matrix obtained from the encoder part of convolution:
\begin{eqnarray}\label{eq:coef0}
C =  \Phi^\top \left( f \circledast  \overline\Psi\right) \ .
 \end{eqnarray}
Here, 
the local filters for the encoder and decoder part convolution are given by 
\begin{eqnarray*}
\overline\Psi := \begin{bmatrix} \overline\psi_1 & \cdots & \overline \psi_q \end{bmatrix} \in \Rd^{d\times q} &,&
\tau(\tilde\Psi) :=   \frac{1}{d} \begin{bmatrix}  \tilde \psi_1 \\ \vdots  \\
\tilde \psi_q
\end{bmatrix}\in \Rd^{dq}  \ . 
\end{eqnarray*}
{\color{black}
where $\overline v$ represents a flipped version of a vector $v$.
}
Moreover, 
for a given  matrix input $Z \in \Rd^{n\times p}$, 
we have the following convolutional framelet expansion \cite{ye2017deep}:
\begin{eqnarray}
Z 
&=& \left(\Phi C\right) \circledast \tau(\tilde \Psi) \label{eq:Zout}
\end{eqnarray}
where the framelet coefficients $C$ is given by
\begin{eqnarray}\label{eq:coef}
C &=&  \Phi^\top \left( Z \circledast  \overline\Psi\right)
 \end{eqnarray}
 Here,  the  encoder and decoder  filters are defined by
\begin{eqnarray}\label{eq:tauZ}
\overline\Psi &:=&\begin{bmatrix}   \overline\psi_1^1 & \cdots &   \overline\psi_q^1  \\ \vdots & \ddots & \vdots \\
\overline\psi_1^p & \cdots &  \overline\psi_q^p 
\end{bmatrix}  \in \Rd^{dp \times q} \\
\tau(\tilde\Psi) &:=&  \frac{1}{d} \begin{bmatrix}  \tilde \psi_1^1 & \cdots &  \tilde \psi_1^p  \\ \vdots & \ddots & \vdots \\
\tilde \psi_q^1 & \cdots &  \tilde \psi_q^p 
\end{bmatrix}  \in \Rd^{dq \times p}
\end{eqnarray}
The simple convolutional framelet expansion using  \eqref{eq:frameeq}, \eqref{eq:Zout},  \eqref{eq:coef0} and \eqref{eq:coef} is so powerful
that a CNN with the encoder-decoder architecture emerges from them by inserting the pair
 \eqref{eq:coef} and \eqref{eq:Zout} between the pair \eqref{eq:coef0} and \eqref{eq:frameeq}.
Moreover, a deep CNN training can be interpreted to learn
the basis matrix $\Psi$ for a given basis $\Phi$ such that maximal energy compaction can be achieved.
For more detail, see \cite{ye2017deep}.

In brief, due to the redundancy in Rx-SC domain data,
the associated Hankel matrix is low-ranked, which makes the convolutional framelet coefficients sparse.
This implies that a directly interpolation of Rx-SC domain RF data is feasible using
deep CNN.

\section{Method}
\label{sec:method}

%
%

%


%

US data in the RF domain were acquired by  E-CUBE 12R  US system (Alpinion Co.,  Korea). Real RF data were acquired by a linear array transducer (L3-12H) with the center frequency of 8.48MHz and a convex array transducer (SC1-4H) with the center frequency of 3.2MHz. The sampling frequencies  were 40MHz. We first acquired RF data from nine subjects using  a linear array transducer. 
The size of each Rx-SC planes is 64$\times$384. 
15000 Rx-SC planes of the seven people data sets are randomly selected for generating training data set, and 3000 Rx-SC planes of the another person data sets are randomly selected for generating validation data set. The remaining one person data sets are used as test set. In addition,
we acquired RF data of liver from  one subject using a convex array transducer. 
This data set was used to verify the universality of the algorithm.
The proposed CNN is composed of convolution layers, batch normalization, ReLU and bypass connection with concatenation as shown in Fig.~\ref{fig:methodb}. Specifically, the network consists of 28 convolution layers composed of a batch normalization and a ReLU except for the last convolution layer. The first 27 convolution layers use 3$\times$3 kernel, and the last convolution layer use 1$\times$1 kernel. Four bypass connections with concatenation exist. As input data, we used randomly sub-sampled data at the downsampling ratio of 4. Because our proposed method assumes a dynamic
aperture US system, the  RF data are sub-sampled along the receivers. And the full data of Rx-SC plane was used as the label data. For network training,
the number of epoch was 400. The regularization parameter was $10^{-4}$. The network was trained by the stochastic gradient descent. The learning rate started from $10^{-7}$ and gradually decreased down to $10^{-9}$. The network was implemented using MatConvNet. 
%
%

\begin{figure}[htb]
  \centering
  \centerline{\includegraphics[width=7.cm,height=8.5cm]{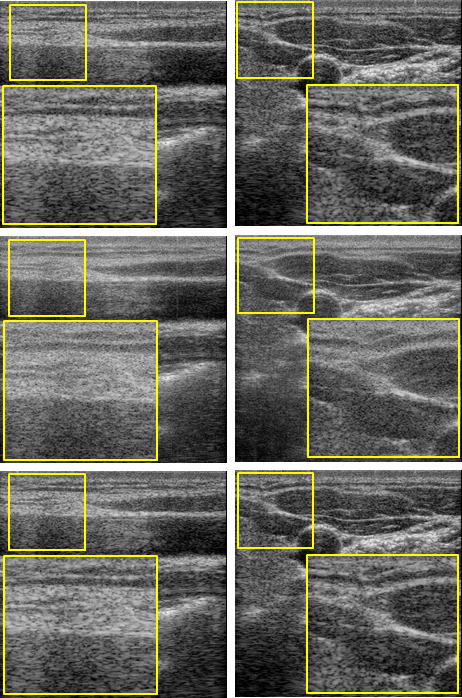}}
  \vspace{-.3cm}
\caption{Linear transducer US images from (first row) full 
RF data, (second row) x4 downsampled RF data,
and (last row) proposed CNN-based interpolation from x4 sub-sampled RF data.
}
\label{fig:resultsb}
\end{figure}


\begin{figure}[htb]
  \centering
  \centerline{\includegraphics[width=8.5cm]{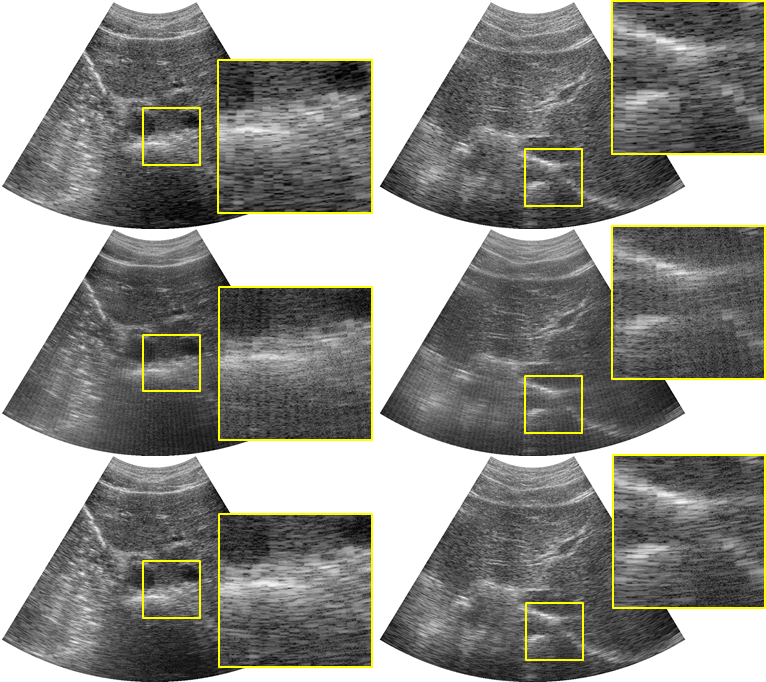}}
\caption{Convex transducer US images from (first row) full 
RF data, (second row) x4 downsampled RF data,
and (last row) proposed CNN-based interpolation from x4 sub-sampled RF data.}
\label{fig:resultsd}
\end{figure}

\section{Experimental Results}
\label{sec:Results}


Fig.~\ref{fig:resultsb} illustrates B-mode ultrasound images acquired from linear array transducer. The proposed CNN-based
interpolation successfully reconstructed the missing RF data from only 25\% of RF data such that
the beamformer removes the blurring artifacts and generates near artifact-free images.
The average of PSNR for the reconstruction B-mode image is around 30.92, which is about 10dB improvement compared to the
B-mode images from sub-sampled RF data.


Next,  we use the network trained using the RF data from  linear array transducer to 
interpolate the missing RF data from a convex array probe. The reconstruction results from a DAS beamformer
are shown in  in Fig.\ref{fig:resultsd}. 
 Because the Rx-SC data of linear array transducer and convex array transducer are similar,
 the DAS beamformer provided very accurate reconstruction results without any line artifacts or blurring from CNN-based RF interpolation.
It is also remarkable that  the accurate reconstruction was obtained for the liver region,
which was never seen by the network trained using the linear array transducer data.
The results confirmed the universality of the algorithm.

\section{Conclusions}
\label{sec:conclusions}

In this paper, we proposed a novel deep learning approach for accelerated B-mode ultrasound imaging.
To exploit the redundancy in the RF domain, the proposed CNN was applied to Rx-SC domain.
Compared with the methods that need to change the hardware, the proposed method does not need any hardware change and can be applied to any B-node ultrasound system or any transducer. 
Therefore, this method can be an important platform for accelerated US imaging.


\end{document}